\documentclass[10pt,twocolumn,letterpaper]{article}

\usepackage{cvpr}
\usepackage{times}
\usepackage{epsfig}
\usepackage{graphicx}
\usepackage{amsmath}
\usepackage{amssymb}

\usepackage[breaklinks=true,bookmarks=false]{hyperref}

\cvprfinalcopy 

\def\cvprPaperID{****} 

\begin{document}

\title{Assessing Architectural Similarity in Populations of Deep Neural Networks}

\author{Audrey G. Chung$^{*}$, Paul Fieguth, and Alexander Wong\\
Vision and Image Processing Research Group, University of Waterloo, Waterloo, ON, Canada \\
Waterloo Artificial Intelligence Institute, University of Waterloo, Waterloo, ON, Canada \\
$^{*}${\tt\small agchung@uwaterloo.ca}
}

\maketitle

\begin{abstract}
   \textbf{Evolutionary deep intelligence} has recently shown great promise for producing small, powerful deep neural network models via the synthesis of increasingly efficient architectures over successive generations. Despite recent research showing the efficacy of multi-parent evolutionary synthesis, little has been done to directly assess architectural similarity between networks during the synthesis process for improved parent network selection. In this work, we present a preliminary study into quantifying architectural similarity via the percentage overlap of architectural clusters. Results show that networks synthesized using architectural alignment (via gene tagging) maintain higher architectural similarities within each generation, potentially restricting the search space of highly efficient network architectures.
\end{abstract}

\section{Introduction}
The use of deep neural networks (DNNs)~\cite{Bengio2009,Lecun2015} has become ubiquitous over the last few years due to their demonstrated efficacy in many challenging application areas, including image classification~\cite{Krizhevsky2012,Simonyan2014}, pose estimation~\cite{Tompson2014,Newell2016}, and speech recognition~\cite{Graves2013,Xiong2018}. However, the modelling accuracy of high-performance DNNs is a result of increased model size and complexity, rendering them impractical for real-world scenarios with limited computational and memory resources. As a result, methods for reducing the computational requirements of DNNs while maintaining performance accuracy are highly desirable.

One such method is \textit{evolutionary deep intelligence}~\cite{Shafiee2018}. Inspired by nature, Shafiee \etal proposed a biologically-motivated method for synthesizing increasingly efficient and compact network architectures over successive generations from existing high-performance DNNs. While the seminal papers in evolutionary deep intelligence~\cite{Shafiee2018,Shafiee2016} formulated the synthesis process as asexual evolutionary synthesis, recent work~\cite{Chung2017_Mating,Chung2017_Poly} has investigated the use of sexual evolutionary synthesis to produce populations of increasingly compact DNNs at each generations.

Most recently, Chung \etal~\cite{Chung2018} conducted an initial study into mitigating architectural mismatch during sexual evolutionary synthesis via a gene tagging system. While results showed no notable difference in performance accuracy, it raises an interesting question: how can we assess the architectural similarity of DNNs in a meaningful and useful way?

In this work, we present a preliminary study exploring the quantification of network architectural similarity in populations of evolutionary synthesized neural networks via percentage overlap of architectural clusters. Architectural similarity is explored within the context of multi-parent sexual evolutionary synthesis, and will allow for the development of improved similarity-based mating policies during the evolutionary synthesis of highly efficient networks.

\section{Methods}
In this paper, we investigate the quantification of architectural similarity using generations of networks synthesized via multi-parent evolutionary synthesis with and without gene tagging~\cite{Chung2018}.

\subsection{$m$-Parent Evolutionary Synthesis}
Let the network architecture be formulated as $\mathcal{H}(N,S)$, where $N$ is the set of possible neurons and $S$ denotes the set of possible synapses in the network. Each neuron $n_j \in N$ is connected to neuron $n_k \in N$ via a set of synapses $\bar{s} \subset S$, such that the synaptic connectivity $s_j \in S$ has an associated $w_j \in W$ to denote the connection's strength. In the seminal evolutionary deep intelligence paper~\cite{Shafiee2016}, the synthesis probability $P(\mathcal{H}_{g}|\mathcal{H}_{g-1}, \mathcal{R}_{g})$ of a new network at generation $g$ is approximated by the synaptic probability $P(S_g|W_{g-1}, R_{g})$ to emulate heredity through the generations of networks. $P(\mathcal{H}_{g}|\mathcal{H}_{g-1}, \mathcal{R}_{g})$ is also conditional on an environmental factor model $\mathcal{R}_g$ to imitate natural selection via simulated environmental resources.

Extending on~\cite{Shafiee2018,Shafiee2016}, Chung~\etal~\cite{Chung2017_Mating} generalized the synthesis process multi-parent ($m$-parent) evolutionary synthesis where a newly synthesized network $\mathcal{H}_{g(i)}$ can be dependent on a subset of all previously synthesized networks $\mathcal{H}_{G_i}$, with $G_i$ corresponding to the set of previous networks on which $\mathcal{H}_{g(i)}$ is dependent and $g(i)$ representing the generation number corresponding to the $i^{\text{th}}$ network.

The synthesis probability combining the probabilities of $m$ parent networks $\mathcal{H}_{G_i}$ is represented by some cluster-level mating function $\mathcal{M}_c(\cdot)$ and some synapse-level mating function $\mathcal{M}_s(\cdot)$:
\begin{align}
P(\mathcal{H}_{g(i)}|\mathcal{H}_{G_i}, \mathcal{R}_{g(i)}) =& \prod_{C \in \mathcal{C}} \Big[ P(s_{g(i),C}|\mathcal{M}_c(W_{\mathcal{H}_{G_i}}), \mathcal{R}_{g(i)}^c) \cdot \nonumber \\ &\prod_{j \in C} P(s_{g(i),j}|\mathcal{M}_s(w_{\mathcal{H}_{G_i},j}), \mathcal{R}_{g(i)}^s) \Big].
\end{align}

\subsection{Architecture Alignment via Gene Tagging}
To encourage like-with-like mating during evolutionary synthesis, Chung \etal~\cite{Chung2018} recently introduced a gene tagging system to enforce structural alignment, \textit{i.e.}, only mating architectural clusters originating from the same location in the ancestor network. As such, the cluster-level and synapse-level mating functions are formulated as follows:
\begin{align}
	\mathcal{M}_c(\overline{W}_{\mathcal{H}_{G_i}}) = \prod_{k \in \mathcal{K}_c} \alpha_{c,k} \overline{W}_{\mathcal{H}_k} \\	\mathcal{M}_s(\overline{w}_{\mathcal{H}_{G_i},j}) = \prod_{k \in \mathcal{K}_c}\alpha_{s,k} \overline{w}_{\mathcal{H}_k,j}
\end{align}
where $\mathcal{K}_c$ is the subset of parent networks with existing architectural clusters corresponding to a single gene tagged cluster $c \in C$, $C$ is the set of clusters that exists in $\mathcal{H}_{g(i)}$, and $\overline{W}$ and $\overline{w}$ are the gene tagged synaptic strengths.

\subsection{Architectural Cluster Overlap}
To investigate the quantification of architectural similarity in the context of multi-parent sexual evolutionary synthesis, the percentage overlap of architectural clusters between two networks is formulated as the proportion of intersecting clusters:
\begin{align}
    \% overlap_{AB} = \frac{C_A \cap C_B}{C_A},
\end{align}
where $C_A$ and $C_B$ represent the sets of architectural clusters that exist in the two networks being compared.

Percentage overlap of architectural clusters is an intuitive representation of network architecture similarity made viable in the context of multi-parent evolutionary synthesis by leveraging the gene tagging system~\cite{Chung2018}. As such, gene tagging (which allows for architectural alignment during evolutionary synthesis) can similarly be used to calculate percentage overlap of existing architectural clusters originating from the same location in the ancestor network. Percentage overlap is indicative of network population diversity within a generation, \eg, relatively low average percentage overlap would indicate a generation of synthesized networks with comparatively higher architectural variability.

\section{Results}
\subsection{Experimental Setup}
In this study, we used the network architectures synthesized in~\cite{Chung2018} with the least aggressive environmental factor model (${R}_{g(i)}^c,{R}_{g(i)}^s= 50$) and trained on the MNIST dataset~\cite{Lecun1998_MNIST}. Architectural similarity was assessed on the first seven generations of networks (after which the performance accuracy degraded to random guessing) synthesized with and without gene tagging.

\subsection{Experimental Results}
\begin{figure}[t]
\begin{center}
  \includegraphics[width=\linewidth]{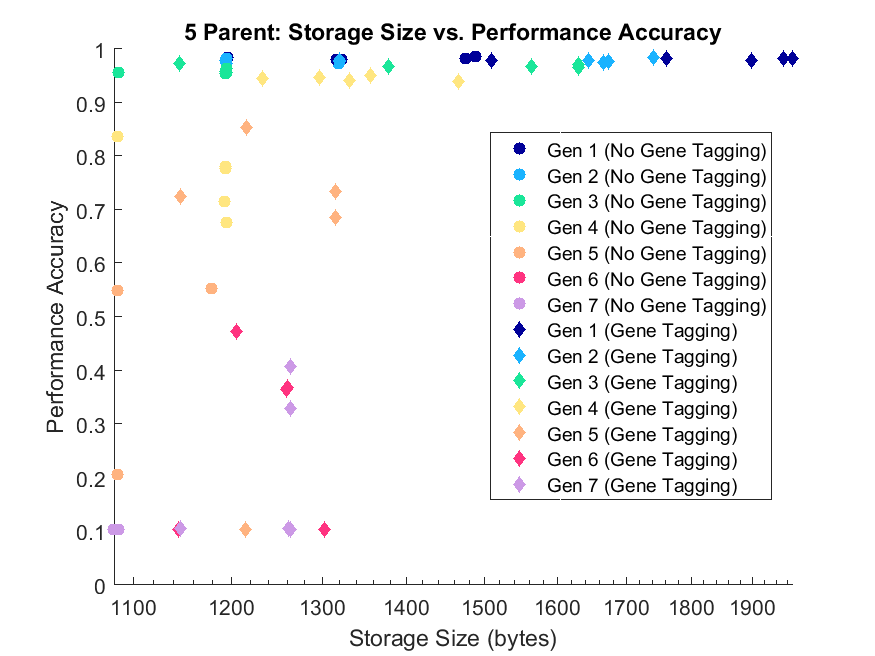}
\end{center}
  \caption{Performance accuracy as a function of storage size for the first seven generations of 5-parent sexual evolutionary synthesis for networks synthesized with gene tagging (diamond) and without gene tagging (round). Plots best viewed in colour.} \label{fig_scatter5}
\end{figure}

Figure~\ref{fig_scatter5} shows the performance accuracy as a function of storage size for the populations of synthesized networks in the first seven generations, where the best synthesized networks are closest to the top left, \textit{i.e.}, high performance accuracy and low storage size. Networks synthesized using gene tagging show a slightly slower progression in maintaining performance accuracy while decreasing storage size relative to networks synthesized without gene tagging.

Synthesizing networks with gene tagging and without gene tagging both produced architectures that increase in variability over successive generations; however, networks synthesized with gene tagging diversify more slowly than those without gene tagging (as shown in Table~\ref{tab_avgOverlap}). Figure~\ref{fig_scatter5} and Table~\ref{tab_avgOverlap} also suggest that generations of networks approaching an optimal tradeoff between performance accuracy and storage size tend to also have the highest architectural variability, \eg, in generations 3 and 4.

Lastly, it is worth noting that the increasing percentage overlap in generations 6 and 7 of networks synthesized without gene tagging is a result of sparse, low-variability architectures that can no longer represent the problem space (\textit{i.e.}, performance accuracy of $10\%$ on the 10-class MNIST dataset, equivalent to random guessing). Similarly, the percentage overlap in generations 6 and 7 of networks synthesized with gene tagging increases as the performance accuracy begins to rapidly decrease.

\section{Discussion}
We presented a preliminary study in assessing architectural similarity between deep neural networks to improve the sexual evolutionary synthesis process. Results show that networks synthesized using gene tagging have less architectural variability than networks synthesized without gene tagging, as quantified by relatively higher overlap percentages of architectural clusters. This indicates that the use of gene tagging is potentially restricting the exploration of highly efficient network architectures in the search space. Future work includes further investigation into quantities of information, \eg, mutual information, as well as the development of a custom similarity metric for optimal architectural similarity during sexual evolutionary synthesis.

\begin{table}[t]
\centering
\caption{Average percentage overlap of architectural clusters in network models for the first seven generations of 5-parent sexual evolutionary synthesis. Note that the increasing percentage overlap in generations 6 and 7 of networks synthesized without gene tagging is a result of sparse, low-variability architectures that can no longer represent the problem space, while the unpredictability of percentage overlap in generations 6 and 7 of networks synthesized with gene tagging may be a result of some (but not all) networks having sparse, low-variability architectures.}
\vspace{10pt}
\begin{tabular}{|c|c|c|}
\hline
\textbf{Gen No.}    & \textbf{Gene Tagging}     & \textbf{No Gene Tagging} \\ \hline \hline
1   & $93.75\%$        &  $93.71\%$   \\ \hline
2   & $87.59\%$        &  $78.11\%$   \\ \hline
3   & $83.49\%$        &  $68.84\%$   \\ \hline
4   & $71.81\%$        &  $66.64\%$   \\ \hline
5   & $73.17\%$        &  $68.44\%$   \\ \hline
6   & $69.09\%$        &  $82.74\%$   \\ \hline
7   & $73.48\%$        &  $91.05\%$   \\ \hline
\end{tabular}
\label{tab_avgOverlap}
\end{table}

{\small
\bibliographystyle{ieee}
\bibliography{EvoNetMI}

\begin{thebibliography}{10}\itemsep=-1pt

\bibitem{Bengio2009}
Y.~Bengio et~al.
\newblock Learning deep architectures for ai.
\newblock {\em Foundations and trends{\textregistered} in Machine Learning},
  2(1):1--127, 2009.

\bibitem{Chung2017_Poly}
A.~G. Chung, P.~Fieguth, and A.~Wong.
\newblock Polyploidism in deep neural networks: m-parent evolutionary synthesis
  of deep neural networks in varying population sizes.
\newblock {\em Journal of Computational Vision and Imaging Systems}, 3(1),
  2017.

\bibitem{Chung2018}
A.~G. Chung, P.~Fieguth, and A.~Wong.
\newblock Mitigating architectural mismatch during the evolutionary synthesis
  of deep neural networks.
\newblock {\em arXiv preprint arXiv:1811.07966}, 2018.

\bibitem{Chung2017_Mating}
A.~G. Chung, M.~Javad~Shafiee, P.~Fieguth, and A.~Wong.
\newblock The mating rituals of deep neural networks: Learning compact feature
  representations through sexual evolutionary synthesis.
\newblock In {\em Proceedings of the IEEE International Conference on Computer
  Vision}, pages 1220--1227, 2017.

\bibitem{Graves2013}
A.~Graves, A.-r. Mohamed, and G.~Hinton.
\newblock Speech recognition with deep recurrent neural networks.
\newblock In {\em 2013 IEEE international conference on acoustics, speech and
  signal processing}, pages 6645--6649. IEEE, 2013.

\bibitem{Krizhevsky2012}
A.~Krizhevsky, I.~Sutskever, and G.~E. Hinton.
\newblock Imagenet classification with deep convolutional neural networks.
\newblock In {\em Advances in neural information processing systems}, pages
  1097--1105, 2012.

\bibitem{Lecun2015}
Y.~LeCun, Y.~Bengio, and G.~Hinton.
\newblock Deep learning.
\newblock {\em nature}, 521(7553):436, 2015.

\bibitem{Lecun1998_MNIST}
Y.~LeCun, C.~Cortes, and C.~Burges.
\newblock {MNIST} handwritten digit database.
\newblock 1998.

\bibitem{Newell2016}
A.~Newell, K.~Yang, and J.~Deng.
\newblock Stacked hourglass networks for human pose estimation.
\newblock In {\em European Conference on Computer Vision}, pages 483--499.
  Springer, 2016.

\bibitem{Shafiee2018}
M.~J. Shafiee, A.~Mishra, and A.~Wong.
\newblock Deep learning with darwin: evolutionary synthesis of deep neural
  networks.
\newblock {\em Neural Processing Letters}, 48(1):603--613, 2018.

\bibitem{Shafiee2016}
M.~J. Shafiee and A.~Wong.
\newblock Evolutionary synthesis of deep neural networks via synaptic
  cluster-driven genetic encoding.
\newblock {\em Advances in Neural Information Processing Systems}, 2016.

\bibitem{Simonyan2014}
K.~Simonyan and A.~Zisserman.
\newblock Very deep convolutional networks for large-scale image recognition.
\newblock {\em arXiv preprint arXiv:1409.1556}, 2014.

\bibitem{Tompson2014}
J.~J. Tompson, A.~Jain, Y.~LeCun, and C.~Bregler.
\newblock Joint training of a convolutional network and a graphical model for
  human pose estimation.
\newblock In {\em Advances in neural information processing systems}, pages
  1799--1807, 2014.

\bibitem{Xiong2018}
W.~Xiong, L.~Wu, F.~Alleva, J.~Droppo, X.~Huang, and A.~Stolcke.
\newblock The microsoft 2017 conversational speech recognition system.
\newblock In {\em 2018 IEEE International Conference on Acoustics, Speech and
  Signal Processing (ICASSP)}, pages 5934--5938. IEEE, 2018.

\end{thebibliography}
}

\end{document}